\documentclass{article}

\usepackage[preprint]{neurips_2026}

\usepackage[utf8]{inputenc} % allow utf-8 input
\usepackage[T1]{fontenc}    % use 8-bit T1 fonts
\usepackage{hyperref}       % hyperlinks
\usepackage{url}            % simple URL typesetting
\usepackage{booktabs}       % professional-quality tables
\usepackage{amsfonts}       % blackboard math symbols
\usepackage{nicefrac}       % compact symbols for 1/2, etc.
\usepackage{microtype}      % microtypography
\usepackage{xcolor}         % colors
\usepackage{graphicx}
\usepackage{multirow} 
\usepackage{amsmath}
\title{Cognitive-Causal Multi-Task Learning with Psychological State Conditioning for Assistive Driving Perception}

\author{%
  Keito Inoshita \\
  Kansai University\\
  3-3-35, Yamatecho, Suita\\
  Osaka, Japan 564-8680 \\
  \texttt{inosita.2865@gmail.com} \\
  \And
  Nobuhiro Hayashida \\
  ISUZU Advanced Engineering Center, Ltd.\\
  8, Tsuchidana, Fujisawa City\\
  Kanagawa, Japan 252-0881 \\
  \texttt{nobuhiro\_hayashida@isuzu.com} \\
  \And
  Akira Imanishi \\
  ISUZU Advanced Engineering Center, Ltd.\\
  8, Tsuchidana, Fujisawa City\\
  Kanagawa, Japan 252-0881 \\
  \texttt{akira\_imanishi@isuzu.com} \\
}

\begin{document}

\maketitle

\begin{abstract}
Multi-task learning for advanced driver assistance systems requires modeling the complex interplay between driver internal states and external traffic environments. However, existing methods treat recognition tasks as flat and independent objectives, failing to exploit the cognitive causal structure underlying driving behavior. In this paper, we propose CauPsi, a cognitive science-grounded causal multi-task learning framework that explicitly models the hierarchical dependencies among Traffic Context Recognition (TCR), Vehicle Context Recognition (VCR), Driver Emotion Recognition (DER), and Driver Behavior Recognition (DBR). The proposed framework introduces two key mechanisms. First, a Causal Task Chain propagates upstream task predictions to downstream tasks via learnable prototype embeddings, realizing the cognitive cascade from environmental perception to behavioral regulation in a differentiable manner. Second, Cross-Task Psychological Conditioning (CTPC) estimates a psychological state signal from driver facial expressions and body posture and injects it as a conditioning input to all tasks including environmental recognition, thereby modeling the modulatory effect of driver internal states on cognitive and decision-making processes. Evaluated on the AIDE dataset, CauPsi achieves a mean accuracy of 82.71\% with only 5.05M parameters, surpassing prior work by +1.0\% overall, with notable improvements on DER (+3.65\%) and DBR (+7.53\%). Ablation studies validate the independent contribution of each component, and analysis of the psychological state signal confirms that it acquires systematic task-label-dependent patterns in a self-supervised manner without explicit psychological annotations.
\end{abstract}

\section{Introduction}

During driving, drivers simultaneously perceive the external traffic environment and regulate their own emotions and behaviors. According to Endsley's situation awareness model \citep{1}, this cognitive processing comprises perception, comprehension, and projection of future states, and is further modulated by the driver's psychological state. As the Yerkes-Dodson law \citep{2} demonstrates, arousal exerts a nonlinear influence on task performance, and even under identical traffic conditions, environmental perception and vehicle operation decisions differ substantially depending on whether the driver is fatigued or tense \citep{3}.

Advanced Driver Assistance Systems (ADAS) have increasingly adopted Multi-Task Learning (MTL) to jointly address Driver Emotion Recognition (DER), Driver Behavior Recognition (DBR), Traffic Context Recognition (TCR), and Vehicle Context Recognition (VCR) \citep{4, 5, 6}. MTL enhances generalization through inter-task feature sharing \citep{7}, yet negative transfer caused by inter-task conflicts remains a challenge \citep{8}. Existing methods have addressed this primarily through feature fusion \citep{9, 10}, but share two fundamental limitations.

First, despite a cognitively well-established causal structure among the four tasks---drivers perceive the traffic situation (TCR), make vehicle operation decisions (VCR), experience emotions through cognitive appraisal (DER), and exhibit behaviors governed by action readiness (DBR)---existing methods treat tasks in a flat manner, failing to exploit the cognitive cascade of ``perception $\to$ judgment $\to$ emotion $\to$ behavior.''

Second, the modulatory effect of driver psychological states on environmental recognition has not been modeled. Driver fatigue degrades cognitive processing capacity, and arousal level directly affects the perceptual accuracy of traffic environments \citep{3}, yet no existing ADAS-oriented MTL method utilizes driver internal states, inferable from facial expressions and body posture, as inputs to TCR or VCR.

To address these challenges, we propose CauPsi, a cognitive science-grounded causal MTL framework comprising: a Causal Task Chain that implements inter-task causal structure as soft-label propagation via prototype embeddings; Cross-Task Psychological Conditioning (CTPC), which estimates a psychological state signal $\boldsymbol{\psi}$ from driver facial expressions and body posture and injects it into all tasks; and bidirectional Cross-View Attention between inside and scene views, built on a frozen MobileNetV3-Small \citep{13} backbone. Our main contributions are:

\begin{itemize}
\item[i)] A Causal Task Chain that realizes the cognitive cascade of environmental perception $\to$ vehicle operation judgment $\to$ emotion elicitation $\to$ behavioral regulation as end-to-end differentiable soft-label propagation via learnable prototype embeddings.

\item[ii)] CTPC, the first framework to incorporate the modulatory effect of driver internal states on environmental cognition into MTL, injecting the psychological state signal $\boldsymbol{\psi}$ estimated from facial expressions and body posture as a conditioning input to all tasks including TCR and VCR.

\item[iii)] CauPsi achieves 82.7\% mean accuracy with only 5.05M parameters on the AIDE benchmark, surpassing prior work with notable improvements on DER (+3.7\%) and DBR (+7.5\%), with ablation studies validating each component's independent contribution.
\end{itemize}
% =====================================================
\section{Related Work}
\label{sec:related}
% =====================================================

\subsection{Multi-Task Learning for Driver Assistance}

MTL improves the generalization performance of individual tasks by sharing representations across tasks \citep{7, 14}. Hard parameter sharing reserves only the final layers for task-specific heads while sharing most parameters across tasks \citep{15, 16}. \citet{17} employed this approach to jointly learn traffic object detection, drivable area segmentation, and lane detection, and \citet{18} realized MTL for panoptic driving perception using an anchor-free architecture. While computationally efficient, hard parameter sharing is susceptible to negative transfer when inter-task discrepancy is large \citep{8}. Soft parameter sharing allows each task to maintain independent parameters while leveraging shared features \citep{19}, as in AdaMV-MoE \citep{21} and task-adaptive attention generators \citep{20}, though at the cost of increased parameter count.

For the four-task ADAS setting on the AIDE dataset \citep{6}, \citet{9} proposed MMTL-UniAD, separating task-shared and task-specific features via dual-branch multimodal embedding, while \citet{10} proposed TEM3-Learning with Mamba-based spatiotemporal feature extraction and gating-based modality fusion, achieving high accuracy across all four tasks under 6M parameters. \citet{22} addressed joint learning of driver emotion and behavior, though without integration of traffic environment recognition tasks. However, all of these methods treat tasks in a flat manner, limiting inter-task information transfer to implicit feature sharing. In contrast, CauPsi incorporates the cognitively grounded causal structure among tasks directly into the network architecture via differentiable soft-label propagation through prototype embeddings.

\subsection{Multimodal Learning and Feature Fusion}

In ADAS, leveraging multiple modalities enables more comprehensive environmental understanding \citep{23, 24}. For driver state recognition, \citet{25} predicted driver behavior by fusing forward-view images, driver images, and vehicle speed data, while \citet{26} improved emotion recognition through a hybrid attention mechanism combining driver images and eye movement data. \citet{27} proposed GLMDriveNet, demonstrating the effectiveness of fusing global and local multimodal features for driving behavior classification.

A key challenge lies in effective inter-modality fusion. Many methods employ independent feature extraction branches per modality \citep{28, 25}, risking the omission of latent inter-modality interactions. \citet{10} addressed this via shared feature extraction across inside-view and scene-view images, while \citet{9} introduced per-task gating to dynamically adjust modality importance. However, these methods perform fusion in a later integration layer without explicit inter-view interaction at the encoder output level. In this work, bidirectional Cross-View Attention between the inside view and scene views explicitly models the interaction between driver internal states and the external environment at the encoder output level.

\subsection{Cognitive, Emotional, and Behavioral Interactions in Driving}
\label{sec:related_cognition}

Cognitive, emotional, and behavioral processes in driving form a mutually interacting hierarchical system. Endsley's SA model \citep{1} formalizes cognitive processing into three levels: perception (Level~1), comprehension (Level~2), and projection (Level~3). \citet{29} demonstrated that SA forms the core of driver decision-making, with TCR corresponding to Level~1--2 and VCR to Level~2--3, establishing a hierarchical dependency between them.

Lazarus's cognitive appraisal theory \citep{30} explains how environmental cognition gives rise to emotion: emotions are elicited by the subject's appraisal of stimuli relative to personal goals, not by external stimuli per se \citep{31}. In driving, recognizing a traffic jam induces anxiety through appraisal of ``delayed arrival,'' while an approaching vehicle triggers fear as a ``threat to safety,'' establishing a causal link from TCR/VCR outputs to DER. Frijda's action tendency theory \citep{12} further links emotion to behavior, anger promotes aggressive driving while fear induces avoidance, establishing a causal link from DER to DBR.

Crucially, this causal chain is not unidirectional: the Yerkes-Dodson law \citep{2} shows that arousal nonlinearly modulates cognitive performance, with fatigue impairing attention and excessive tension inducing attentional narrowing \citep{32}. Russell's circumplex model \citep{11} provides a unified two-dimensional (arousal--valence) framework for quantifying these psychological states. Together, these theories establish the causal structure TCR/VCR $\to$ DER $\to$ DBR with psychological state modulation at every stage. Nevertheless, existing ADAS-oriented MTL methods have not incorporated these insights: although face and body information is used for DER/DBR, no mechanism feeds back the estimated psychological state into TCR or VCR. CTPC addresses this gap by estimating $\boldsymbol{\psi}$ from facial expressions and body posture and injecting it as a conditioning input to all tasks.
% =====================================================
\section{Cognitive-Causal Multi-Task Learning with Psychological State Conditioning}
\label{sec:method}
% =====================================================

\subsection{Framework Overview}

\begin{figure*}[t]
\centering
\includegraphics[width=\textwidth]{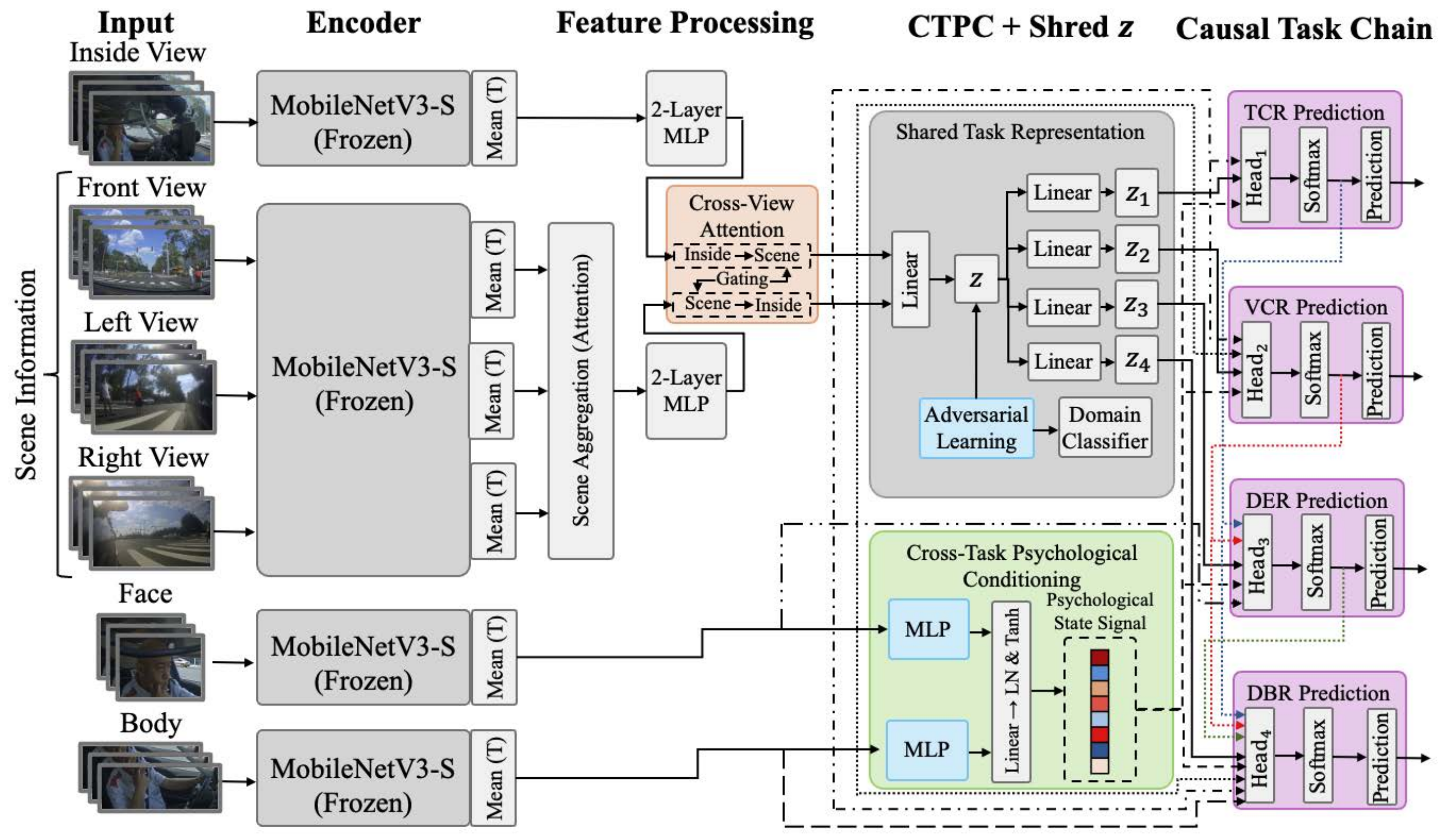}
\caption{Overall architecture of CauPsi.}
\label{fig:architecture}
\end{figure*}

CauPsi jointly recognizes driver cognitive, emotional, and behavioral states alongside the traffic environment from multi-view video. As shown in Figure.~\ref{fig:architecture}, the input comprises an inside view, multiple scene views, and cropped face/body regions, each provided as $T$ consecutive frames. The framework consists of: i) multi-view feature processing with bidirectional Cross-View Attention between the inside and scene views; ii) CTPC, which estimates the psychological state signal $\boldsymbol{\psi}$ from driver facial expressions and body posture and injects it as a conditioning input to all tasks; iii) a Causal Task Chain that explicitly models inter-task causal dependencies via prototype embeddings; and iv) loss functions and training stabilization techniques.

The overall processing flow is as follows. Each view's video is processed by a frozen pre-trained encoder and temporally aggregated via average pooling, after which scene view features are integrated through an attention mechanism. Bidirectional Cross-View Attention is then applied between the inside-view and scene-view features, and the fused representation is linearly projected to obtain the latent representation $\mathbf{z}$. In parallel, CTPC estimates $\boldsymbol{\psi}$ from face and body features. Finally, the Causal Task Chain produces hierarchical predictions for all four tasks, taking as input $\mathbf{z}$, individual view features, prototype embeddings from upstream tasks, and $\boldsymbol{\psi}$.

% ---------------------------------------------------
\subsection{Multi-View Feature Processing with Cross-View Attention}
\label{sec:feature_extraction}

Let the input for view $v$ be $\mathbf{X}_v \in \mathbb{R}^{C \times T \times H_v \times W_v}$. Frames are processed by frozen pre-trained encoders $\phi_{\mathrm{in}}$, $\phi_{\mathrm{sc}}$ (shared across scene views), $\phi_{\mathrm{face}}$, and $\phi_{\mathrm{body}}$, followed by Global Average Pooling (GAP) and temporal averaging:
\begin{equation}
\bar{\mathbf{h}}_v = \frac{1}{T} \sum_{t=1}^{T} \left(\mathbf{W}_v^{\mathrm{gap}} \cdot \mathrm{GAP}\!\left(\phi_v\!\left(\mathbf{X}_v^{(t)}\right)\right) + \mathbf{b}_v^{\mathrm{gap}}\right) \in \mathbb{R}^{d_c}
\end{equation}
Face and body features $\mathbf{f}_{\mathrm{face}}, \mathbf{f}_{\mathrm{body}} \in \mathbb{R}^{d_f}$ are obtained analogously with a nonlinear projection. The $N_s$ scene view features are aggregated via attention-weighted fusion:
\begin{equation}
\bar{\mathbf{h}}_{\mathrm{scene}} = \sum_{i=1}^{N_s} \alpha_i \, \bar{\mathbf{h}}_{\mathrm{sc},i}, \quad
\boldsymbol{\alpha} = \mathrm{softmax}\!\left(\mathbf{W}_2 \cdot \mathrm{ReLU}\!\left(\mathbf{W}_1 [\bar{\mathbf{h}}_{\mathrm{sc},1}; \cdots; \bar{\mathbf{h}}_{\mathrm{sc},N_s}]\right)\right)
\end{equation}
The attention weights $\boldsymbol{\alpha}$ dynamically adjust the contribution of each viewpoint according to the scene context. The inside-view and scene features are projected to $\mathbf{f}_{\mathrm{in}}, \mathbf{f}_{\mathrm{scene}} \in \mathbb{R}^{d_f}$ via two-layer MLPs.

To capture the interaction between driver internal states and the external environment, bidirectional Cross-View Attention is applied between $\mathbf{f}_{\mathrm{in}}$ and $\mathbf{f}_{\mathrm{scene}}$. For the Inside$\to$Scene direction:

\begin{align}
\mathbf{c}_{\mathrm{in}} &= \mathrm{MHA}(Q\!=\!\mathrm{LN}(\mathbf{f}_{\mathrm{in}}),\; K\!=\!\mathbf{f}_{\mathrm{scene}},\; V\!=\!\mathbf{f}_{\mathrm{scene}}) \\
\mathbf{g}_{\mathrm{in}} &= \sigma\!\left(\mathbf{W}_g^{\mathrm{in}}[\mathbf{f}_{\mathrm{in}}; \mathbf{c}_{\mathrm{in}}] + \mathbf{b}_g^{\mathrm{in}}\right) \\
\tilde{\mathbf{f}}_{\mathrm{in}} &= \mathbf{f}_{\mathrm{in}} + \mathbf{g}_{\mathrm{in}} \odot \mathbf{c}_{\mathrm{in}}
\end{align}

The Scene$\to$Inside direction is defined symmetrically, yielding $\tilde{\mathbf{f}}_{\mathrm{scene}}$. The gating mechanism $\mathbf{g}$ adaptively controls the amount of cross-view information incorporated at each dimension, and applying Layer Normalization (LN) only to the query ensures stable attention computation.

% ---------------------------------------------------
\subsection{Cross-Task Psychological Conditioning}
\label{sec:ctpc}

Driver psychological states modulate environmental cognition and behavioral decision-making, yet this effect has not been modeled in existing methods. CTPC estimates $\boldsymbol{\psi}$ from face and body features and injects it as a conditioning input to all task predictions. Motivated by Russell's circumplex model \citep{11} and Frijda's action tendency theory \citep{12}, CTPC introduces a structural inductive bias: affect-related components are extracted from $\mathbf{f}_{\mathrm{face}}$ and action-related components from $\mathbf{f}_{\mathrm{body}}$ via independent two-layer MLPs:
\begin{equation}
\mathbf{a}_{\mathrm{affect}} = \mathbf{W}_a^{(2)} \cdot \mathrm{ReLU}\!\left(\mathbf{W}_a^{(1)} \mathbf{f}_{\mathrm{face}} + \mathbf{b}_a^{(1)}\right) + \mathbf{b}_a^{(2)} \in \mathbb{R}^{d_\psi}
\end{equation}
\begin{equation}
\mathbf{a}_{\mathrm{action}} = \mathbf{W}_r^{(2)} \cdot \mathrm{ReLU}\!\left(\mathbf{W}_r^{(1)} \mathbf{f}_{\mathrm{body}} + \mathbf{b}_r^{(1)}\right) + \mathbf{b}_r^{(2)} \in \mathbb{R}^{d_\psi}
\end{equation}
\begin{equation}
\boldsymbol{\psi} = \tanh\!\left(\mathrm{LN}\!\left(\mathbf{W}_\psi [\mathbf{a}_{\mathrm{affect}}; \mathbf{a}_{\mathrm{action}}] + \mathbf{b}_\psi\right)\right) \in \mathbb{R}^{d_\psi}
\end{equation}
The Tanh activation constrains $\boldsymbol{\psi} \in [-1, 1]$, enabling representation of psychological polarity (positive/negative valence, high/low arousal). Critically, $\boldsymbol{\psi}$ requires no ground-truth psychological labels and is learned entirely via backpropagation from the four task losses, acquiring driver internal state representations in a self-supervised manner. The structurally separated pathways for face and body provide affect-related and action-related inductive biases, but what information is ultimately encoded in $\boldsymbol{\psi}$ is determined by the task losses. Whereas prior methods use face/body information only for DER/DBR, CTPC injects $\boldsymbol{\psi}$ into all tasks including TCR and VCR, incorporating the Yerkes-Dodson modulatory effect \citep{2} directly into the architecture.

% ---------------------------------------------------
\subsection{Causal Task Chain}
\label{sec:causal_chain}

\begin{figure*}[t]
\centering
\includegraphics[width=0.55\textwidth]{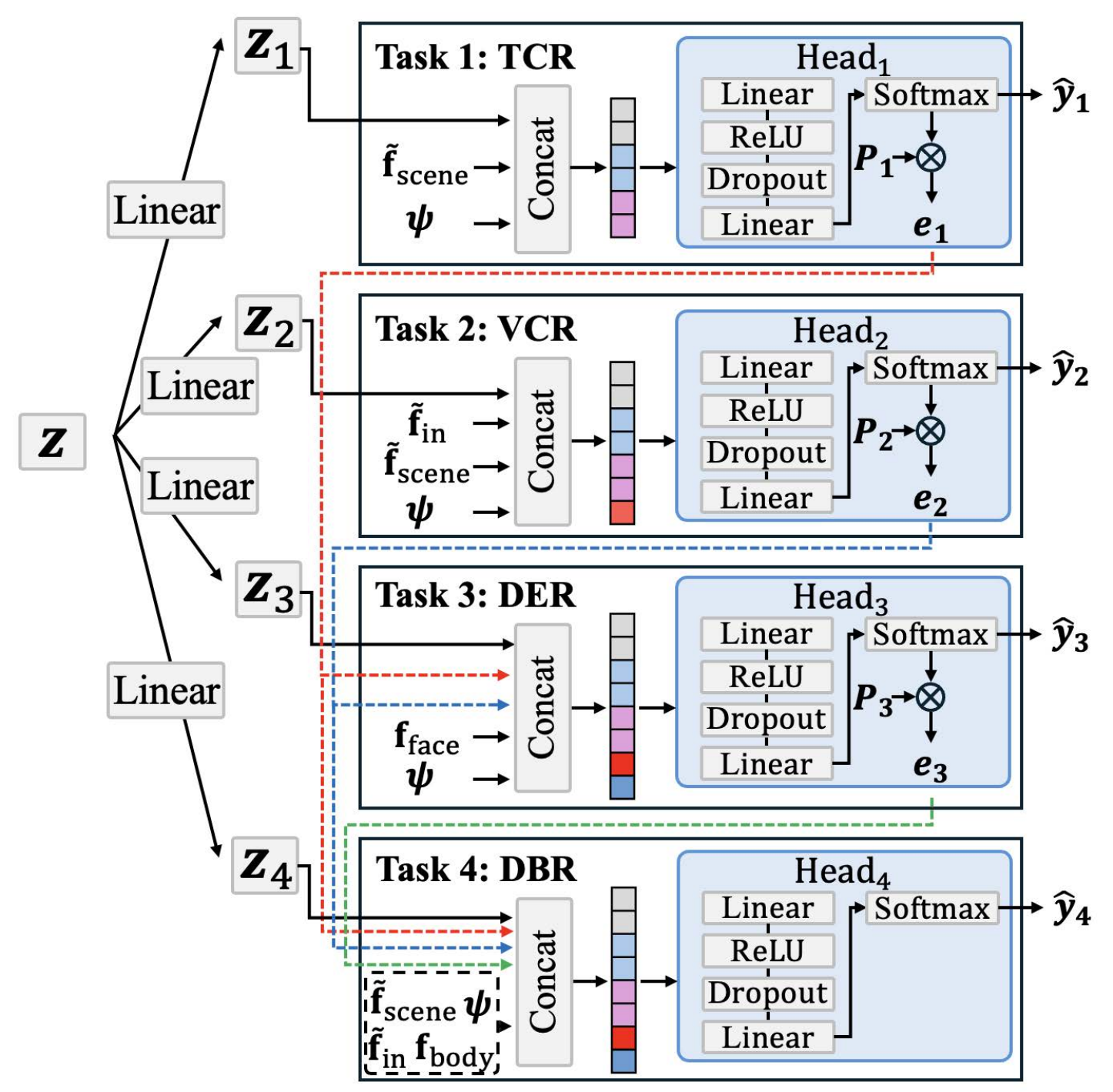}
\caption{Detailed architecture of the Causal Task Chain.}
\label{fig:causal_chain}
\end{figure*}

A causal structure of environmental cognition $\to$ vehicle operation judgment $\to$ emotion elicitation $\to$ behavioral regulation exists among the four recognition tasks. The Causal Task Chain explicitly models this structure by propagating upstream task predictions to downstream tasks as learnable prototype embeddings, as illustrated in Figure.~\ref{fig:causal_chain}. The Cross-View Attention-enhanced features are projected to task-shared and task-specific representations:
\begin{equation}
\mathbf{z} = \mathbf{W}_z [\tilde{\mathbf{f}}_{\mathrm{in}}; \tilde{\mathbf{f}}_{\mathrm{scene}}] + \mathbf{b}_z \in \mathbb{R}^{d_z}, \qquad
\mathbf{z}_r = \mathbf{W}_{\pi_r} \mathbf{z} + \mathbf{b}_{\pi_r} \in \mathbb{R}^{d_t}
\end{equation}
For upstream tasks ($r = 1, 2, 3$), the soft-label embedding $\mathbf{e}_r = \hat{\mathbf{y}}_r \cdot \mathbf{P}_r \in \mathbb{R}^{d_e}$ is computed from the predicted distribution $\hat{\mathbf{y}}_r$ and a learnable prototype matrix $\mathbf{P}_r \in \mathbb{R}^{C_r \times d_e}$, and propagated to downstream tasks. Since $\hat{\mathbf{y}}_r$ is softmax-normalized, $\mathbf{e}_r$ is a confidence-weighted average of per-class prototype vectors. Unlike hard-label propagation via argmax, this is fully differentiable, enabling gradients from downstream losses to backpropagate through upstream parameters and realizing end-to-end cooperative learning.

The four tasks are predicted hierarchically, with each task head being a two-layer MLP ($\mathrm{Linear} \to \mathrm{ReLU} \to \mathrm{Dropout} \to \mathrm{Linear}$) with softmax output. The input composition reflects the progressive accumulation of cognitive information per Endsley's SA model \citep{1}:

\begin{align}
\hat{\mathbf{y}}_1 &= \mathrm{Head}_1([\mathbf{z}_1;\, \tilde{\mathbf{f}}_{\mathrm{scene}};\, \boldsymbol{\psi}]) \quad \text{(TCR)} \\
\hat{\mathbf{y}}_2 &= \mathrm{Head}_2([\mathbf{z}_2;\, \tilde{\mathbf{f}}_{\mathrm{in}};\, \tilde{\mathbf{f}}_{\mathrm{scene}};\, \boldsymbol{\psi}]) \quad \text{(VCR)} \\
\hat{\mathbf{y}}_3 &= \mathrm{Head}_3([\mathbf{z}_3;\, \mathbf{e}_1;\, \mathbf{e}_2;\, \mathbf{f}_{\mathrm{face}};\, \boldsymbol{\psi}]) \quad \text{(DER)} \\
\hat{\mathbf{y}}_4 &= \mathrm{Head}_4([\mathbf{z}_4;\, \mathbf{e}_3;\, \mathbf{e}_1;\, \mathbf{e}_2;\, \tilde{\mathbf{f}}_{\mathrm{scene}};\, \tilde{\mathbf{f}}_{\mathrm{in}};\, \mathbf{f}_{\mathrm{body}};\, \boldsymbol{\psi}]) \quad \text{(DBR)}
\end{align}

TCR receives only scene features as the most upstream task; VCR additionally receives inside-view features; DER receives upstream embeddings $\mathbf{e}_1, \mathbf{e}_2$ grounded in Lazarus's appraisal theory \citep{30}; DBR integrates all upstream embeddings alongside multi-view and body features per Frijda's action tendency theory \citep{12}, reflecting that behavior emerges as the cumulative outcome of all preceding cognitive stages.

% ---------------------------------------------------
\subsection{Loss Functions and Training}
\label{sec:loss}

Each task is trained with class-weighted Label Smoothing Cross-Entropy loss, with higher weights assigned to psychological state-related tasks ($\lambda_3 > \lambda_1$, $\lambda_4 > \lambda_2$) to promote learning on tasks with pronounced class imbalance:
\begin{equation}
\mathcal{L} = \sum_{r=1}^{4} \lambda_r \, \mathcal{L}_{\mathrm{CE}}^{(r)}(\hat{\mathbf{y}}_r, y_r;\, \mathbf{w}_r, \epsilon) + \gamma_{\mathrm{adv}} \, \mathcal{L}_{\mathrm{adv}}
\end{equation}
The adversarial loss $\mathcal{L}_{\mathrm{adv}}$ is produced by a domain classifier with a Gradient Reversal Layer (GRL), regularizing $\mathbf{z}$ to suppress domain-specific information via unsupervised K-means domain labels. Training is further stabilized via frozen encoders, Exponential Moving Average (EMA) of model parameters, mixup augmentation, gradient accumulation, and warmup cosine annealing. The complete processing procedure of CauPsi is summarized in Appendix~\ref{sec:appendix_algorithm}, and full training details including the EMA and learning rate schedule are provided in Appendix~\ref{sec:appendix_training}.
% =====================================================
\section{Experiments}
\label{sec:experiments}
% =====================================================

\subsection{Experimental Setup}

\textbf{Dataset.} We evaluate CauPsi on the AIDE dataset \citep{6}, an open-source multimodal time-series dataset for driver assistance research comprising 2,898 samples. Each sample includes multi-view images from four viewpoints (front, left, right, and inside) and is annotated with labels for all four tasks: DER, DBR, TCR, and VCR. The dataset is split into training, validation, and test sets at ratios of 65\%, 15\%, and 20\%, respectively. As preprocessing, the driver's face and upper-body regions are cropped from inside-view images based on bounding box coordinates. Each input sequence consists of 16 consecutive frames at 16 fps, and random horizontal flipping is applied as data augmentation.

\textbf{Implementation Details.} MobileNetV3-Small \citep{13} pre-trained on ImageNet is used as the frozen backbone encoder. Optimization uses AdamW (learning rate $3 \times 10^{-4}$, weight decay $1 \times 10^{-4}$) with Warmup Cosine Annealing. The effective batch size is 64 via gradient accumulation over 4 steps. Task loss weights are set to $\lambda_{\mathrm{TCR}} = 1.0$, $\lambda_{\mathrm{VCR}} = 1.0$, $\lambda_{\mathrm{DER}} = 1.5$, $\lambda_{\mathrm{DBR}} = 2.0$, assigning higher weights to psychological state-related tasks. Full hyperparameter details are provided in Appendix~\ref{sec:appendix_impl}.

\textbf{Metrics.} Per-task accuracy ($\alpha_{\mathrm{acc}}$) and mean accuracy across all four tasks ($\beta_{\mathrm{macc}}$) are the primary evaluation metrics. Macro-averaged F1 score is additionally reported as a supplementary metric.

% ---------------------------------------------------
\subsection{Comparison with Existing Methods}
\label{sec:comparison}

Table~\ref{tab:comparison} presents a comparison with existing methods following the experimental setup of \citet{10}. Methods are categorized by backbone type: 2D Convolutional Neural Network (CNN), 2D CNN with temporal embedding, and 3D CNN. CauPsi achieves a mean accuracy of 82.71\%, surpassing TEM$^3$-Learning \citep{10}, the highest-performing and most parameter-efficient prior model, by 1.0 percentage point (81.68\%). Most notably, substantial improvements are observed on DER (78.65\%, $+$3.65\%) and DBR (76.84\%, $+$7.53\%), both directly linked to driver psychological states, suggesting that the Causal Task Chain and CTPC function effectively for psychological state-related tasks.

\begin{table*}[t]
\centering
\caption{Comparison with existing methods.}
\label{tab:comparison}
\resizebox{\textwidth}{!}{%
\begin{tabular}{l|lll|cccc|c|c}
\toprule
\multirow{2}{*}{Pattern} & \multicolumn{3}{c|}{Backbone} & \multicolumn{4}{c|}{$\alpha_{\mathrm{acc}}$ (\%) $\uparrow$} & $\beta_{\mathrm{macc}}$ & P(M) \\
 & Multi-view Scene & Driver Images & Joints & DER & DBR & TCR & VCR & (\%) $\uparrow$ & $\downarrow$ \\
\midrule
\multirow{4}{*}{2D}
 & VGG16 \citep{33} & VGG16 \citep{33} & 3DCNN & 69.12 & 64.57 & 84.77 & 74.08 & 73.15 & 127.48 \\
 & Res18 \citep{34} & Res18 \citep{34} & 3DCNN & 68.78 & 64.33 & 89.76 & 78.59 & 75.37 & 107.77 \\
 & CMT \citep{35} & CMT \citep{35} & 3DCNN & 68.75 & 68.75 & 93.75 & 81.38 & 78.16 & 72.33 \\
 & GLMDriveNet \citep{27} & GLMDriveNet \citep{27} & 3DCNN & 71.38 & 66.57 & 90.23 & 77.19 & 76.34 & 78.17 \\
\midrule
\multirow{5}{*}{\shortstack{2D +\\Timing}}
 & PP-Res18+TransE & Res18/34+TransE & MLP+TE & 70.83 & 67.32 & 90.54 & 79.97 & 77.17 & - \\
 & Res34+TransE & Res18/34+TransE & MLP+TE & 72.65 & 67.08 & 86.63 & 78.46 & 76.21 & - \\
 & Res50+TransE & Res34/50+TransE & MLP+TE & 70.24 & 65.65 & 82.57 & 77.29 & 73.94 & - \\
 & VGG16+TransE & VGG13/16+TransE & MLP+TE & 71.12 & 67.15 & 85.13 & 78.58 & 75.50 & - \\
 & VGG19+TransE & VGG16/19+TransE & MLP+TE & 69.46 & 65.48 & 85.74 & 77.91 & 74.65 & - \\
\midrule
\multirow{11}{*}{3D}
 & 3D-Res34 \citep{36} & 3D-Res34 \citep{36} & 3DCNN & 69.13 & 63.05 & 87.82 & 79.31 & 74.83 & 303.10 \\
 & MobileNet-V1-3D \citep{13} & MobileNet-V1-3D \citep{13} & ST-GCN & 72.23 & 64.20 & 88.34 & 77.83 & 75.65 & 54.05 \\
 & MobileNet-V2-3D \citep{37} & MobileNet-V2-3D \citep{37} & ST-GCN & 68.47 & 61.74 & 86.54 & 78.66 & 73.85 & 83.78 \\
 & ShuffleNet-V1-3D \citep{38} & ShuffleNet-V1-3D \citep{38} & ST-GCN & 72.41 & 68.97 & 90.64 & 80.79 & 78.20 & 31.49 \\
 & ShuffleNet-V2-3D \citep{39} & ShuffleNet-V2-3D \citep{39} & ST-GCN & 70.94 & 64.04 & 89.33 & 78.98 & 75.82 & 35.09 \\
 & C3D \citep{40} & C3D \citep{40} & ST-GCN & 63.05 & 63.95 & 85.41 & 77.01 & 72.36 & 158.46 \\
 & I3D \citep{41} & I3D \citep{41} & ST-GCN & 70.94 & 66.17 & 87.68 & 79.81 & 76.15 & - \\
 & SlowFast \citep{42} & SlowFast \citep{42} & ST-GCN & 72.38 & 61.58 & 86.86 & 78.33 & 74.79 & - \\
 & TimeSFormer \citep{43} & TimeSFormer \citep{43} & ST-GCN & 74.87 & 65.18 & 92.12 & 78.81 & 77.75 & 158.46 \\
 & Video Swin Trans. \citep{44} & Video Swin Trans. \citep{44} & 3DCNN & 73.44 & 65.63 & 93.75 & 75.00 & 76.96 & 119.80 \\
 & MARNet \citep{9} & MARNet \citep{9} & 3DCNN & 76.67 & 73.61 & 93.19 & 85.00 & 82.30 & - \\
 & MTS-Mamba \citep{10} & MTS-Mamba \citep{10} & 3DCNN & 75.00 & 69.31 & \textbf{96.29} & \textbf{86.11} & 81.68 & 5.99 \\ 

\midrule
CauPsi (ours) & MobileNetV3-S & MobileNetV3-S & --- & \textbf{78.65} & \textbf{76.84} & 92.11 & 83.25 & \textbf{82.71} & \textbf{5.05} \\
\bottomrule
\end{tabular}
}
\end{table*}

CauPsi falls below TEM$^3$-Learning on TCR (92.11\%, $-$4.18\%) and VCR (83.25\%, $-$2.86\%), attributable to encoder architecture differences: whereas TEM$^3$-Learning directly models spatiotemporal features via a Mamba-based State Space Model, CauPsi processes each frame independently with MobileNetV3-Small and aggregates via temporal averaging, limiting dynamic temporal modeling. Nevertheless, CauPsi achieves this with only 5.05M parameters, a 16\% reduction from TEM$^3$-Learning's 5.99M, demonstrating a competitive balance between efficiency and accuracy. Furthermore, CauPsi uses no joint position information, yet surpasses methods that incorporate joint data in mean accuracy, confirming that face and body features combined with CTPC serve as an effective alternative representation of driver state. Per-class analysis is provided in Appendix~\ref{sec:appendix_perclass}.

% ---------------------------------------------------
\subsection{Ablation Study}
\label{sec:ablation}

To quantitatively evaluate the contribution of each component, we design five ablation conditions, each removing one component while keeping all training settings and random seeds fixed. Results are presented in Table~\ref{tab:ablation}. The full model outperforms all ablation conditions. Removing face and body features causes the largest drop ($-$2.46\%), with DBR declining 6.73 points (76.84\%$\to$70.11\%), demonstrating that localized face and body features are indispensable for behavior recognition. A 3.12-point drop on VCR further confirms that fine-grained driver appearance contributes to vehicle operation judgment as well. 

\begin{table}[t]
\centering
\caption{Ablation study results. $\Delta$ denotes the change in mean accuracy relative to the full model.}
\label{tab:ablation}
\small
\begin{tabular}{l|cc|cccc}
\toprule
Model & $\beta_{\mathrm{macc}}$ & $\Delta$ & DER & DBR & TCR & VCR \\
\midrule
CauPsi & \textbf{82.71} & --- & 78.65 & 76.84 & 92.11 & 83.25 \\
$-$CTPC & 81.77 & $-$0.94 & 79.63 & 73.89 & 91.95 & 81.60 \\
$-$CrossView & 82.38 & $-$0.33 & 79.96 & 75.69 & 92.28 & 81.60 \\
$-$CausalChain & 80.41 & $-$2.30 & 78.16 & 68.14 & 92.93 & 82.43 \\
$-$FaceBody & 80.25 & $-$2.46 & 78.65 & 70.11 & 92.11 & 80.13 \\
\bottomrule
\end{tabular}
\end{table}

Removing the Causal Task Chain ($-$2.30\%) yields the largest single-task drop: DBR declines 8.70 points (76.84\%$\to$68.14\%), strongly indicating that prototype embedding-based causal propagation from upstream tasks is decisive for behavior recognition. Without the Causal Task Chain, each task predicts solely from its task-specific projection $\mathbf{z}_r$ and scene features, unable to exploit upstream cognitive outputs. Removing CTPC ($-$0.94\%) degrades VCR ($-$1.65\%) and DBR ($-$2.95\%), corroborating that $\boldsymbol{\psi}$ contributes to both environmental and behavioral recognition. Cross-View Attention contributes modestly overall ($-$0.33\%) but shows a $-$1.65\% effect on VCR. The full model's superiority is consistent on $\beta_{\mathrm{macc}}$, reflecting that per-task trade-offs are inherent to MTL.

% ---------------------------------------------------
\subsection{Analysis of the Psychological State Signal}
\label{sec:psi_analysis}

\begin{figure*}[t]
\centering
\includegraphics[width=0.9\textwidth]{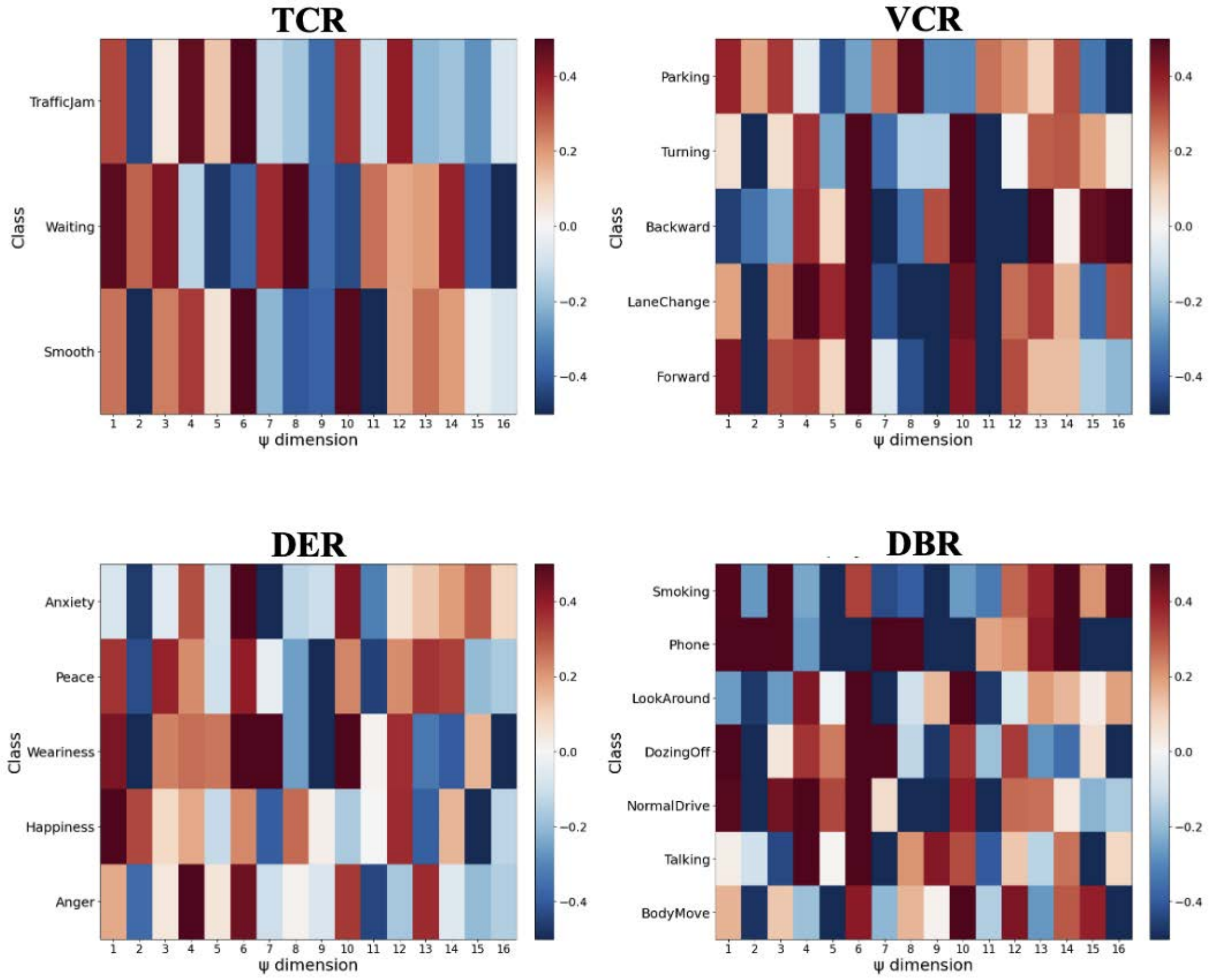}
\caption{Mean values across the 16 dimensions of $\boldsymbol{\psi}$, stratified by class for each task.}
\label{fig:psi_heatmap}
\end{figure*}

To verify whether $\boldsymbol{\psi}$ acquires meaningful representations that vary systematically with task labels, we collect $\boldsymbol{\psi}$ from all 609 test-set samples. Figure.~\ref{fig:psi_heatmap} presents heatmaps of the mean values across all 16 dimensions, stratified by class for each task.

Clear inter-class differences in $\boldsymbol{\psi}$ patterns are observed across all tasks. In DER, Weariness and Happiness form contrasting patterns on d7 and d10, consistent with opposing poles of the arousal axis in Russell's circumplex model \citep{11}. In DBR, Phone and DozingOff are encoded by qualitatively distinct dimensional patterns, reflecting the difference between active attentional distraction and passive arousal degradation. For VCR, LaneChange shows the largest positive value on d6 among all classes, possibly reflecting the higher cognitive load of lane changing relative to other maneuvers.

Cross-task analysis reveals that d7 consistently takes positive values for low-arousal passive states (Weariness, DozingOff) and negative values for high-arousal externally directed states (Anxiety, LookAround, Backward), suggesting it encodes arousal-related information corresponding to the affect-related component in CTPC. Crucially, these systematic patterns are acquired without any explicit psychological state annotations, through minimization of task losses alone, confirming that CTPC extracts meaningful driver internal state representations in a self-supervised manner. Detailed per-dimension analysis is provided in Appendix~\ref{sec:appendix_psi}.
% =====================================================
\section{Discussion}
\label{sec:discussion}
% =====================================================

\subsection{Key Findings}

The most important finding of this work is that the Causal Task Chain, which propagates upstream task predictions to downstream tasks via prototype embeddings, is indispensable for behavior recognition. Its removal causes DBR to decline sharply by 8.70 percentage points (76.84\%$\to$68.14\%), the largest single-task drop across all ablation conditions. This result underscores a broader insight: driver emotion and behavior cannot be treated in isolation from the vehicle and traffic context. Just as Frijda's action tendency theory \citep{12} holds that behavior emerges as the cumulative outcome of cognition and emotion, a driver's emotional and behavioral states are inherently shaped by the surrounding traffic situation. Modeling driver and vehicle context jointly, rather than independently, is therefore essential for accurate recognition. Soft-label propagation via prototype embeddings realizes this joint modeling in a differentiable manner, enabling upstream task training to also contribute to downstream task accuracy.

A second key finding is the empirical demonstration that the psychological state signal $\boldsymbol{\psi}$, extracted from driver facial expressions and body posture, enables face and body information previously used exclusively for DER and DBR to be leveraged for TCR and VCR as well. The CTPC ablation results in a mean accuracy drop of 0.94 percentage points, with pronounced effects on VCR ($-$1.65\%) and DBR ($-$2.95\%), consistent with the Yerkes-Dodson law \citep{2} that vehicle operation judgment varies with driver internal state even under identical traffic conditions. While it is well established in cognitive psychology that arousal level and fatigue affect environmental cognition performance \citep{3, 29}, this work is the first to model this effect in an MTL framework and empirically demonstrate a corresponding accuracy improvement.

Notably, face and body features serve a dual function in CauPsi. Removing them disables CTPC as well (effectively $\boldsymbol{\psi} = \mathbf{0}$), so the accuracy drop ($-$2.46\%) subsumes the CTPC contribution ($-$0.94\%). The residual ($-$1.52\%) represents the direct contribution of face and body features to the DER and DBR task heads, independent of CTPC. This confirms that face and body information simultaneously acts as direct local appearance descriptors and as a source of psychological state information mediated by $\boldsymbol{\psi}$.

In contrast, CauPsi falls below TEM$^3$-Learning \citep{10} on TCR ($-$4.18\%) and VCR ($-$2.86\%), attributable to the encoder architecture: frame-level CNN processing with temporal averaging loses dynamic temporal information, as directly evidenced by LaneChange in VCR being misclassified as Forward at a rate of 78\% (Appendix~\ref{sec:appendix_perclass}). The structural strengths of CauPsi---CTPC and the Causal Task Chain---partially compensate for this encoder-level limitation but do not fully overcome it.

\subsection{Limitations}

First, although $\boldsymbol{\psi}$ exhibits systematic task-label-dependent patterns, whether individual dimensions correspond to specific psychological indicators such as arousal, valence, or cognitive load remains unverified. Validation using datasets with ground-truth psychological labels (e.g., physiological arousal ratings) is necessary to establish the interpretability of $\boldsymbol{\psi}$.

Second, frame-level CNN encoding with temporal averaging discards sequential dynamics, directly causing low accuracy on VCR classes requiring motion cues (LaneChange: F1 $= 0.217$, Turning: F1 $= 0.695$). Introducing spatiotemporal approaches such as State Space Models or Temporal Attention is a natural next step.

Third, label smoothing and class weighting are insufficient for severely underrepresented classes (DozingOff: 12 samples, F1 $= 0.727$; Anger: 45 samples, F1 $= 0.518$). Oversampling, stronger data augmentation, or few-shot learning frameworks warrant investigation.
% =====================================================
\section{Conclusion}
\label{sec:conclusion}
% =====================================================

We proposed CauPsi, a cognitive science-grounded causal MTL framework for jointly addressing DER, DBR, TCR, and VCR. The Causal Task Chain implements inter-task causal structure as differentiable soft-label propagation via prototype embeddings, and CTPC estimates the psychological state signal $\boldsymbol{\psi}$ from driver facial expressions and body posture and injects it into all tasks. On the AIDE dataset, CauPsi achieves 82.71\% mean accuracy with 5.05M parameters, surpassing prior work by 1.0 percentage point with notable gains on DER (+3.65\%) and DBR (+7.53\%). This work provides the first empirical demonstration that driver internal states can be incorporated as a modulatory signal in MTL for ADAS, suggesting that inter-task causal dependencies and psychological state conditioning are important design principles for future ADAS-oriented multi-task learning.

\begin{ack}
The authors would like to express their sincere gratitude to Isuzu Motors Limited, Isuzu Advanced Engineering Center Limited, and NineSigma Holdings, Inc. for their valuable advice and insightful feedback throughout this study. The authors also thank Dr. Hirotaka Hara for his technical advice and constructive suggestions.
\end{ack}

\bibliographystyle{abbrvnat}  % NeurIPS標準スタイル
\bibliography{refs}            % refs.bibを参照

@article{1,
  author    = {Endsley, M. R.},
  title     = {Toward a theory of situation awareness in dynamic systems},
  journal   = {Human Factors},
  year      = {1995},
  volume    = {37},
  number    = {1},
  pages     = {32--64},
  doi       = {10.1518/001872095779049543},
}

@article{2,
  author    = {Yerkes, R. M. and Dodson, J. D.},
  title     = {The relation of strength of stimulus to rapidity of habit-formation},
  journal   = {Journal of Comparative Neurology and Psychology},
  year      = {1908},
  volume    = {18},
  number    = {5},
  pages     = {459--482},
  doi       = {10.1002/cne.920180503},
}

@article{3,
  author    = {Liu, Y.-C. and Wu, T.-J.},
  title     = {Fatigued driver's driving behavior and cognitive task performance: Effects of road environments and road environment changes},
  journal   = {Safety Science},
  year      = {2009},
  volume    = {47},
  number    = {8},
  pages     = {1083--1089},
  doi       = {10.1016/j.ssci.2008.11.009},
}

@article{4,
  author    = {Gong, Y. and Lu, J. and Liu, W. and Li, Z. and Jiang, X. and Gao, X. and others},
  title     = {{SIFDriveNet}: Speed and image fusion for driving behavior classification network},
  journal   = {IEEE Transactions on Computational Social Systems},
  year      = {2024},
  volume    = {11},
  number    = {1},
  pages     = {1244--1259},
  doi       = {10.1109/tcss.2023.3303334},
}

@article{5,
  author    = {Zhang, X. and Gong, Y. and Lu, J. and Li, Z. and Li, S. and Wang, S. and others},
  title     = {Oblique convolution: A novel convolution idea for redefining Lane detection},
  journal   = {IEEE Transactions on Intelligent Vehicles},
  year      = {2024},
  volume    = {9},
  number    = {2},
  pages     = {4025--4039},
  doi       = {10.1109/tiv.2023.3319547},
}

@inproceedings{6,
  author    = {Yang, D. and Huang, S. and Xu, Z. and Li, Z. and Wang, S. and Li, M. and others},
  title     = {{AIDE}: A Vision-Driven Multi-View, Multi-Modal, Multi-Tasking Dataset for Assistive Driving Perception},
  booktitle = {Proceedings of the IEEE/CVF International Conference on Computer Vision},
  month     = {October},
  year      = {2023},
  pages     = {20459-20470},
  url       = {http://arxiv.org/abs/2307.13933},
}

@inproceedings{7,
  author    = {Chowdhuri, S. and Pankaj, T. and Zipser, K.},
  title     = {{MultiNet}: Multi-modal multi-task learning for autonomous driving},
  booktitle = {2019 IEEE Winter Conference on Applications of Computer Vision},
  publisher = {IEEE},
  year      = {2019},
  pages     = {1496--1504},
  doi       = {10.1109/wacv.2019.00164},
}

@inproceedings{8,
  author    = {Liu, S. and Liang, Y. and Gitter, A.},
  title     = {Loss-Balanced Task Weighting to reduce negative transfer in multi-task learning},
  booktitle = {Proceedings of the AAAI Conference on Artificial Intelligence},
  year      = {2019},
  volume    = {33},
  number    = {01},
  pages     = {9977--9978},
  doi       = {10.1609/aaai.v33i01.33019977},
}

@misc{9,
  author    = {Liu, W. and Wang, W. and Qiao, Y. and Guo, Q. and Zhu, J. and Li, P. and others},
  title     = {{MMTL-UniAD}: A Unified Framework for Multimodal and Multi-Task Learning in Assistive Driving Perception},
  booktitle = {Proceedings of the 2026 IEEE/CVF Conference on Computer Vision and Pattern Recognition},
  year      = {2026},
  pages     = {6864--6874},
  doi       = {10.48550/ARXIV.2504.02264},
}

@misc{10,
  author    = {Liu, W. and Qiao, Y. and Wang, Z. and Guo, Q. and Chen, Z. and Zhou, M. and others},
  title     = {{TEM\^{}3-learning}: Time-Efficient Multimodal Multi-task learning for advanced assistive driving},
  year      = {2025},
  doi       = {10.48550/ARXIV.2506.18084},
  archivePrefix = {arXiv},
  primaryClass  = {cs.CV},
}

@article{11,
  author    = {Russell, J. A.},
  title     = {A circumplex model of affect},
  journal   = {Journal of Personality and Social Psychology},
  year      = {1980},
  volume    = {39},
  number    = {6},
  pages     = {1161--1178},
  doi       = {10.1037/h0077714},
}

@article{12,
  author    = {Frijda, N. H.},
  title     = {Emotion, cognitive structure, and action tendency},
  journal   = {Cognition \& Emotion},
  year      = {1987},
  volume    = {1},
  number    = {2},
  pages     = {115--143},
  doi       = {10.1080/02699938708408043},
}

@inproceedings{13,
  author    = {Howard, A. and Sandler, M. and Chen, B. and Wang, W. and Chen, L.-C. and Tan, M. and others},
  title     = {Searching for {MobileNetV3}},
  booktitle = {Proceedings of the 2019 IEEE/CVF International Conference on Computer Vision},
  year      = {2019},
  pages     = {1314--1324},
  doi       = {10.1109/iccv.2019.00140},
}

@inproceedings{14,
  author    = {Ishihara, K. and Kanervisto, A. and Miura, J. and Hautamaki, V.},
  title     = {Multi-task learning with attention for end-to-end autonomous driving},
  booktitle = {Proceedings of the 2021 IEEE/CVF Conference on Computer Vision and Pattern Recognition Workshops},
  year      = {2021},
  pages     = {2896--2905},
  doi       = {10.1109/cvprw53098.2021.00325},
}

@inproceedings{15,
  author    = {Cao, K. and You, J. and Leskovec, J.},
  title     = {Relational multi-task learning: Modeling relations between data and tasks},
  booktitle = {Proceedings of the 10th International Conference on Learning Representations},
  year      = {2023},
  doi       = {10.48550/ARXIV.2303.07666},
  pages     = {1--14},
}

@inproceedings{16,
  author    = {Cui, J. and Du, J. and Liu, W. and Lian, Z.},
  title     = {{TextNeRF}: A novel scene-text image synthesis method based on neural radiance fields},
  booktitle = {Proceedings of the 2024 IEEE/CVF Conference on Computer Vision and Pattern Recognition},
  publisher = {IEEE},
  year      = {2024},
  volume    = {34},
  pages     = {22272--22281},
  doi       = {10.1109/cvpr52733.2024.02102},
}

@article{17,
  author    = {Wu, D. and Liao, M.-W. and Zhang, W.-T. and Wang, X.-G. and Bai, X. and Cheng, W.-Q. and others},
  title     = {{YOLOP}: You only look once for panoptic driving perception},
  journal   = {Machine Intelligence Research},
  year      = {2022},
  volume    = {19},
  number    = {6},
  pages     = {550--562},
  doi       = {10.1007/s11633-022-1339-y},
}

@article{18,
  author    = {Zhan, J. and Luo, Y. and Guo, C. and Wu, Y. and Meng, J. and Liu, J.},
  title     = {{YOLOPX}: Anchor-free multi-task learning network for panoptic driving perception},
  journal   = {Pattern Recognition},
  year      = {2024},
  volume    = {148},
  pages     = {110152},
  doi       = {10.1016/j.patcog.2023.110152},
}

@article{19,
  author    = {Gao, M. and Li, J.-Y. and Chen, C.-H. and Li, Y. and Zhang, J. and Zhan, Z.-H.},
  title     = {Enhanced multi-task learning and knowledge graph-based recommender system},
  journal   = {IEEE Transactions on Knowledge and Data Engineering},
  year      = {2023},
  volume    = {35},
  number    = {10},
  pages     = {10281--10294},
  doi       = {10.1109/tkde.2023.3251897},
}

@inproceedings{20,
  author    = {Choi, W. and Shin, M. and Lee, H. and Cho, J. and Park, J. and Im, S.},
  title     = {Multi-task learning for real-time autonomous driving leveraging task-adaptive attention generator},
  booktitle = {Proceedings of the 2024 IEEE International Conference on Robotics and Automation},
  year      = {2024},
  volume    = {2},
  pages     = {14732--14739},
  doi       = {10.1109/icra57147.2024.10610716},
}

@inproceedings{21,
  author    = {Chen, T. and Chen, X. and Du, X. and Rashwan, A. and Yang, F. and Chen, H. and others},
  title     = {{AdaMV-MoE}: Adaptive multi-task vision mixture-of-experts},
  booktitle = {Proceedings of the 2023 IEEE/CVF International Conference on Computer Vision},
  year      = {2023},
  pages     = {17300--17311},
  doi       = {10.1109/iccv51070.2023.01591},
}

@article{22,
  author    = {Xing, Y. and Lv, C. and Cao, D. and Velenis, E.},
  title     = {Multi-scale driver behavior modeling based on deep spatial-temporal representation for intelligent vehicles},
  journal   = {Transportation Research Part C: Emerging Technologies},
  year      = {2021},
  volume    = {130},
  pages     = {103288},
  doi       = {10.1016/j.trc.2021.103288},
}

@article{23,
  author    = {Alaba, S. Y. and Gurbuz, A. C. and Ball, J. E.},
  title     = {Emerging trends in autonomous vehicle perception: Multimodal fusion for {3D} object detection},
  journal   = {World Electric Vehicle Journal},
  year      = {2024},
  volume    = {15},
  number    = {1},
  pages     = {20},
  doi       = {10.3390/wevj15010020},
}

@article{24,
  author    = {Li, Z. and Zhang, T. and Zhou, M. and Tang, D. and Zhang, P. and Liu, W. and others},
  title     = {{MIPD}: A multi-sensory interactive perception dataset for embodied intelligent driving},
  journal   = {IEEE Transactions on Intelligent Transportation Systems},
  year      = {2025},
  volume    = {26},
  number    = {11},
  pages     = {21320--21334},
  doi       = {10.1109/tits.2025.3593298},
}

@article{25,
  author    = {Zhou, D. and Liu, H. and Ma, H. and Wang, X. and Zhang, X. and Dong, Y.},
  title     = {Driving behavior prediction considering cognitive prior and driving context},
  journal   = {IEEE Transactions on Intelligent Transportation Systems},
  year      = {2021},
  volume    = {22},
  number    = {5},
  pages     = {2669--2678},
  doi       = {10.1109/tits.2020.2973751},
}

@article{26,
  author    = {Mou, L. and Zhao, Y. and Zhou, C. and Nakisa, B. and Rastgoo, M. N. and Ma, L. and others},
  title     = {Driver emotion recognition with a hybrid attentional multimodal fusion framework},
  journal   = {IEEE Transactions on Affective Computing},
  year      = {2023},
  volume    = {14},
  number    = {4},
  pages     = {2970--2981},
  doi       = {10.1109/taffc.2023.3250460},
}

@article{27,
  author    = {Liu, W. and Gong, Y. and Zhang, G. and Lu, J. and Zhou, Y. and Liao, J.},
  title     = {{GLMDriveNet}: Global--local multimodal fusion driving behavior classification network},
  journal   = {Engineering Applications of Artificial Intelligence},
  year      = {2024},
  volume    = {129},
  pages     = {107575},
  doi       = {10.1016/j.engappai.2023.107575},
}

@article{28,
  author    = {Guo, C. and Liu, H. and Chen, J. and Ma, H.},
  title     = {Temporal information fusion network for driving behavior prediction},
  journal   = {IEEE Transactions on Intelligent Transportation Systems},
  year      = {2023},
  volume    = {24},
  number    = {9},
  pages     = {9415--9424},
  doi       = {10.1109/tits.2023.3267150},
}

@incollection{29,
  author    = {Baumann, M. and Krems, J. F.},
  title     = {Situation awareness and driving: A cognitive model},
  booktitle = {Modelling Driver Behaviour in Automotive Environments},
  publisher = {Springer London},
  address   = {London},
  year      = {2007},
  pages     = {253--265},
  doi       = {10.1007/978-1-84628-618-6_14},
}

@book{30,
  author    = {Lazarus, R. S.},
  title     = {Emotion and Adaptation},
  publisher = {Oxford University Press},
  address   = {New York, NY},
  year      = {1991},
  doi       = {10.1093/oso/9780195069945.001.0001},
}

@article{31,
  author    = {Moors, A.},
  title     = {On the causal role of appraisal in emotion},
  journal   = {Emotion Review},
  year      = {2013},
  volume    = {5},
  number    = {2},
  pages     = {132--140},
  doi       = {10.1177/1754073912463601},
}

@inproceedings{32,
  author    = {Hadi, F. and Lee, J. Y. and Yeoh, W. L. and Bu, N. and Fukuda, O.},
  title     = {Driving fatigue detection based on behavioral with cognitive task scenario},
  booktitle = {Proceedings of the 2025 10th International Conference on Intelligent Informatics and Biomedical Sciences},
  year      = {2025},
  volume    = {10},
  pages     = {264--269},
  doi       = {10.1109/iciibms66230.2025.11316713},
}

@misc{33,
  author    = {Simonyan, K. and Zisserman, A.},
  title     = {Very deep convolutional networks for large-scale image recognition},
  year      = {2014},
  doi       = {10.48550/arXiv.1409.1556},
  archivePrefix = {arXiv},
  primaryClass  = {cs.CV},
}

@inproceedings{34,
  author    = {He, K. and Zhang, X. and Ren, S. and Sun, J.},
  title     = {Deep Residual Learning for Image Recognition},
  booktitle = {Proceedings of the 2016 IEEE Conference on Computer Vision and Pattern Recognition},
  year      = {2016},
  pages     = {770--778},
  url       = {10.1109/cvpr.2016.90},
}

@inproceedings{35,
  author    = {Guo, J. and Han, K. and Wu, H. and Tang, Y. and Chen, X. and Wang, Y. and others},
  title     = {{CMT}: Convolutional neural networks meet vision transformers},
  booktitle = {Proceedings of the 2022 IEEE/CVF Conference on Computer Vision and Pattern Recognition},
  year      = {2022},
  pages     = {12165--12175},
  doi       = {10.1109/cvpr52688.2022.01186},
}

@inproceedings{36,
  author    = {Hara, K. and Kataoka, H. and Satoh, Y.},
  title     = {Can Spatiotemporal {3D CNNs} Retrace the History of {2D CNNs} and {ImageNet}?},
  booktitle = {Proceedings of the 2018 IEEE/CVF Conference on Computer Vision and Pattern Recognition},
  year      = {2018},
  pages     = {6546--6555},
  doi       = {10.1109/cvpr.2018.00685},
}

@inproceedings{37,
  author    = {Sandler, M. and Howard, A. and Zhu, M. and Zhmoginov, A. and Chen, L.-C.},
  title     = {{MobileNetV2}: Inverted residuals and linear bottlenecks},
  booktitle = {Proceedings of the 2018 IEEE/CVF Conference on Computer Vision and Pattern Recognition},
  year      = {2018},
  pages     = {4510--4520},
  doi       = {10.1109/cvpr.2018.00474},
}

@inproceedings{38,
  author    = {Zhang, X. and Zhou, X. and Lin, M. and Sun, J.},
  title     = {{ShuffleNet}: An extremely efficient convolutional neural network for mobile devices},
  booktitle = {Proceedings of the 2018 IEEE/CVF Conference on Computer Vision and Pattern Recognition},
  year      = {2018},
  pages     = {6848--6856},
  doi       = {10.1109/cvpr.2018.00716},
}

@inproceedings{39,
  author    = {Ma, N. and Zhang, X. and Zheng, H.-T. and Sun, J.},
  title     = {{ShuffleNet V2}: Practical guidelines for efficient {CNN} architecture design},
  booktitle = {Computer Vision -- ECCV 2018},
  series    = {Lecture Notes in Computer Science},
  year      = {2018},
  pages     = {122--138},
  doi       = {10.1007/978-3-030-01264-9_8},
}

@inproceedings{40,
  author    = {Tran, D. and Bourdev, L. and Fergus, R. and Torresani, L. and Paluri, M.},
  title     = {Learning Spatiotemporal Features with {3D} Convolutional Networks},
  booktitle = {Proceedings of the 2015 IEEE International Conference on Computer Vision},
  year      = {2015},
  doi       = {10.1109/iccv.2015.510},
}

@inproceedings{41,
  author    = {Carreira, J. and Zisserman, A.},
  title     = {Quo Vadis, action recognition? A new model and the kinetics dataset},
  booktitle = {Proceedings of the 2017 IEEE Conference on Computer Vision and Pattern Recognition},
  year      = {2017},
  doi       = {10.1109/cvpr.2017.502},
}

@inproceedings{42,
  author    = {Feichtenhofer, C. and Fan, H. and Malik, J. and He, K.},
  title     = {{SlowFast} Networks for Video Recognition},
  booktitle = {Proceedings of the 2019 IEEE/CVF International Conference on Computer Vision},
  year      = {2019},
  doi       = {10.1109/iccv.2019.00630},
}

@inproceedings{43,
  author    = {Bertasius, G. and Wang, H. and Torresani, L.},
  title     = {Is space-time attention all you need for video understanding?},
  year      = {2021},
  booktitle = {Proceedings of the 38th International Conference on Machine
Learning},
  vol       = {139},
  pages     = {813--824},
  url       = {http://arxiv.org/abs/2102.05095},
}

@inproceedings{44,
  author    = {Liu, Z. and Ning, J. and Cao, Y. and Wei, Y. and Zhang, Z. and Lin, S. and others},
  title     = {Video Swin Transformer},
  year      = {2021},
  booktitle = {Proceedings of the 2022 IEEE/CVF Conference on Computer Vision and Pattern Recognition},
  pages     = {3192--3201},
  url       = {10.1109/CVPR52688.2022.00320},
}

% \clearpage  % 参考文献とAppendixの間で改ページ
%%%%%%%%%%%%%%%%%%%%%%%%%%%%%%%%%%%%%%%%%%%%%%%%%%%%%%%%%%%%

\appendix
\section{Overall Algorithm of CauPsi}
\label{sec:appendix_algorithm}

Algorithm 1 summarizes the complete processing procedure of CauPsi.

\begin{figure}[h]
\begin{center}
\small
\begin{tabular}{l}
\hline
Algorithm 1: Overall Algorithm of CauPsi \\
\hline
\textbf{Input:} Multi-view video $\{\mathbf{X}_v\}$, face/body video $\mathbf{X}_{\mathrm{face}}, \mathbf{X}_{\mathrm{body}}$ \\
\textbf{Output:} Four-task predictions $\hat{\mathbf{y}}_1, \hat{\mathbf{y}}_2, \hat{\mathbf{y}}_3, \hat{\mathbf{y}}_4$ \\
\hline
\textit{// Feature Extraction (Section~\ref{sec:feature_extraction})} \\
1: For each view $v$: $\bar{\mathbf{h}}_v \leftarrow \frac{1}{T}\sum_t \mathbf{W}_v^{\mathrm{gap}} \cdot \mathrm{GAP}(\phi_v(\mathbf{X}_v^{(t)}))$ \\
2: $\mathbf{f}_{\mathrm{face}}, \mathbf{f}_{\mathrm{body}} \leftarrow$ face/body encoders $+$ projection \\
3: $\bar{\mathbf{h}}_{\mathrm{scene}} \leftarrow \sum_i \alpha_i \bar{\mathbf{h}}_{\mathrm{sc},i}$ \quad (scene integration) \\
4: $\mathbf{f}_{\mathrm{in}}, \mathbf{f}_{\mathrm{scene}} \leftarrow$ two-layer MLP projection \\
5: $\tilde{\mathbf{f}}_{\mathrm{in}}, \tilde{\mathbf{f}}_{\mathrm{scene}} \leftarrow$ CrossViewAttention \\
\\
\textit{// Psychological State Conditioning (Section~\ref{sec:ctpc})} \\
6: $\mathbf{a}_{\mathrm{affect}} \leftarrow \mathrm{MLP}(\mathbf{f}_{\mathrm{face}})$ \\
7: $\mathbf{a}_{\mathrm{action}} \leftarrow \mathrm{MLP}(\mathbf{f}_{\mathrm{body}})$ \\
8: $\boldsymbol{\psi} \leftarrow \tanh(\mathrm{LN}(\mathbf{W}_\psi[\mathbf{a}_{\mathrm{affect}}; \mathbf{a}_{\mathrm{action}}]))$ \\
\\
\textit{// Task-Shared Representation (Section~\ref{sec:causal_chain})} \\
9: $\mathbf{z} \leftarrow \mathbf{W}_z [\tilde{\mathbf{f}}_{\mathrm{in}}; \tilde{\mathbf{f}}_{\mathrm{scene}}]$ \\
10: $\mathbf{z}_r \leftarrow \mathbf{W}_{\pi_r} \mathbf{z}$ \quad ($r = 1,2,3,4$) \\
\\
\textit{// Causal Task Chain (Section~\ref{sec:causal_chain})} \\
11: $\hat{\mathbf{y}}_1 \leftarrow \mathrm{Head}_1([\mathbf{z}_1; \tilde{\mathbf{f}}_{\mathrm{scene}}; \boldsymbol{\psi}])$,\quad
$\mathbf{e}_1 \leftarrow \hat{\mathbf{y}}_1 \cdot \mathbf{P}_1$ \\
12: $\hat{\mathbf{y}}_2 \leftarrow \mathrm{Head}_2([\mathbf{z}_2; \tilde{\mathbf{f}}_{\mathrm{in}}; \tilde{\mathbf{f}}_{\mathrm{scene}}; \boldsymbol{\psi}])$,\quad
$\mathbf{e}_2 \leftarrow \hat{\mathbf{y}}_2 \cdot \mathbf{P}_2$ \\
13: $\hat{\mathbf{y}}_3 \leftarrow \mathrm{Head}_3([\mathbf{z}_3; \mathbf{e}_1; \mathbf{e}_2; \mathbf{f}_{\mathrm{face}}; \boldsymbol{\psi}])$,\quad
$\mathbf{e}_3 \leftarrow \hat{\mathbf{y}}_3 \cdot \mathbf{P}_3$ \\
14: $\hat{\mathbf{y}}_4 \leftarrow \mathrm{Head}_4([\mathbf{z}_4; \mathbf{e}_3; \mathbf{e}_1; \mathbf{e}_2; \tilde{\mathbf{f}}_{\mathrm{scene}}; \tilde{\mathbf{f}}_{\mathrm{in}}; \mathbf{f}_{\mathrm{body}}; \boldsymbol{\psi}])$ \\
\\
\textit{// Adversarial Training (Section~\ref{sec:loss})} \\
15: $\hat{\mathbf{d}} \leftarrow \mathrm{MLP}(\mathrm{GRL}_\lambda(\mathbf{z}))$ \\
16: \textbf{return} $\hat{\mathbf{y}}_1, \hat{\mathbf{y}}_2, \hat{\mathbf{y}}_3, \hat{\mathbf{y}}_4$ \\
\hline
\end{tabular}
\end{center}
\label{alg:overall}
\end{figure}

\section{Training Details}
\label{sec:appendix_training}

Model parameters are updated with EMA at decay rate $\beta$:
\begin{equation}
\boldsymbol{\theta}_{\mathrm{EMA}}^{(t+1)} = \beta \, \boldsymbol{\theta}_{\mathrm{EMA}}^{(t)} + (1 - \beta) \, \boldsymbol{\theta}^{(t)}
\end{equation}
The EMA parameters are used at evaluation time. The learning rate follows linear warm-up over the first $E_w$ epochs, then decays via cosine annealing to $\eta_{\min}$:
\begin{equation}
\eta(e) = \begin{cases}
\eta_{\max} \cdot \dfrac{e+1}{E_w} & \text{if } e < E_w \\[8pt]
\eta_{\min} + \dfrac{\eta_{\max} - \eta_{\min}}{2}\!\left(1 + \cos\!\left(\dfrac{\pi(e - E_w)}{E - E_w}\right)\right) & \text{otherwise}
\end{cases}
\end{equation}
Gradients are accumulated over $A=4$ steps before each parameter update, with gradient norm clipping applied at each step. Early stopping is triggered when validation accuracy fails to improve for $P$ consecutive epochs, and the best EMA model is used for test evaluation.

\section{Implementation Details}
\label{sec:appendix_impl}

The backbone encoder is MobileNetV3-Small \citep{13} pre-trained on ImageNet, with all parameters frozen during training. Optimization uses AdamW (lr $3\times10^{-4}$, weight decay $1\times10^{-4}$) with Warmup Cosine Annealing (5 warm-up epochs, $\eta_{\min} = 1\times10^{-6}$). Batch size is 16 with gradient accumulation over $A=4$ steps (effective batch size 64). EMA decay $\beta=0.999$, Mixup $\alpha_{\mathrm{mix}}=0.2$, gradient clipping max norm 5.0, early stopping patience 20, maximum 100 epochs. The label smoothing parameter is $\epsilon=0.1$ and the adversarial loss weight is $\gamma_{\mathrm{adv}}=0.5$. Domain labels are generated via K-means clustering with optimal $K$ selected by the Silhouette Score. Model hyperparameters: $d_f=128$, $d_z=256$, $d_t=64$, $d_e=32$, $d_\psi=16$, Cross-View Attention heads $=4$.

\section{Per-Class Analysis}
\label{sec:appendix_perclass}

\begin{figure*}[h]
\centering
\includegraphics[width=\textwidth]{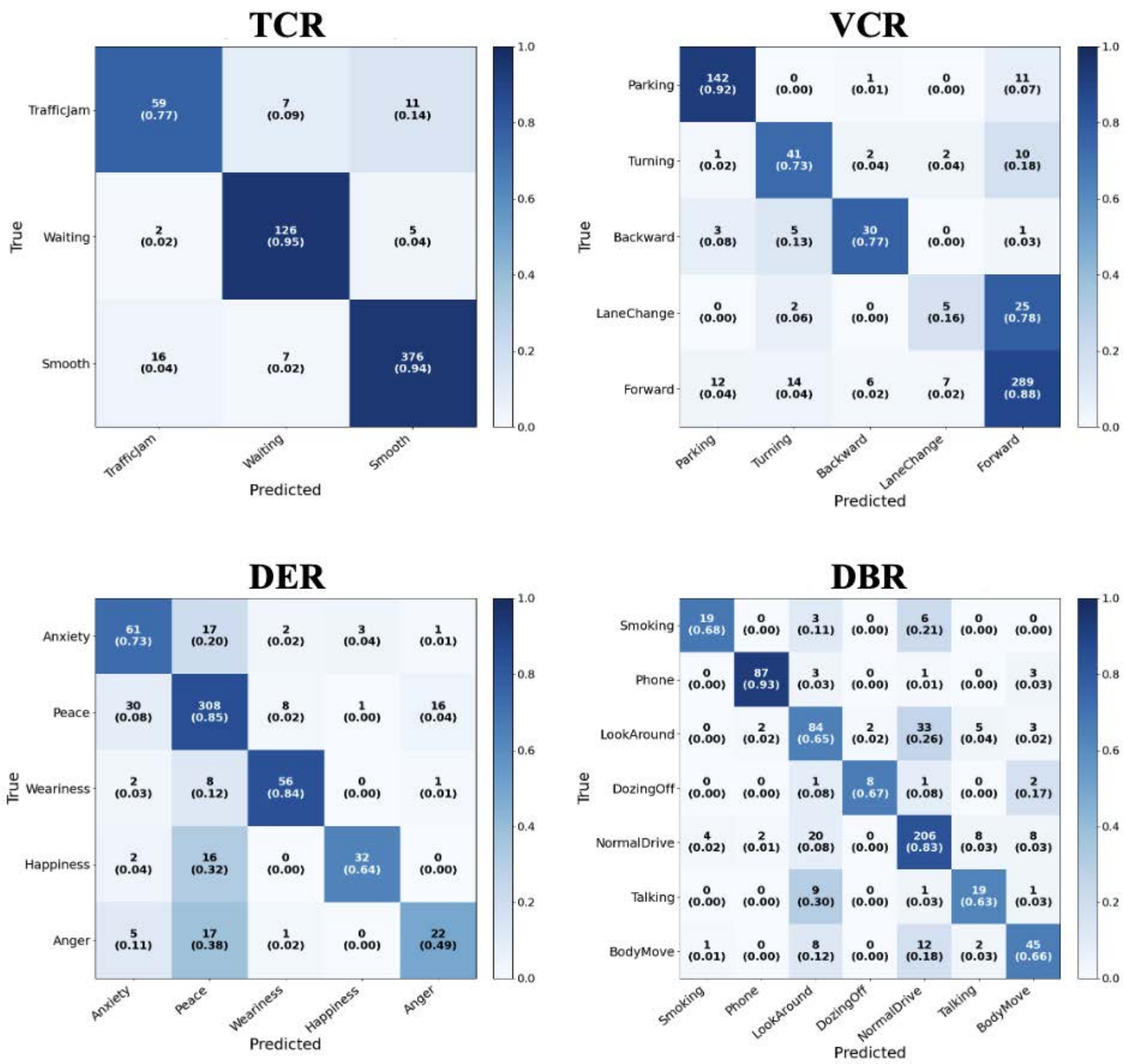}
\caption{Normalized confusion matrices for all four tasks.}
\label{fig:confusion}
\end{figure*}

Figure.~\ref{fig:confusion} presents normalized confusion matrices for all four tasks, and Table~\ref{tab:per_class} reports per-class Precision, Recall, and F1 scores. For TCR, Waiting attains the highest F1 (0.923) and Smooth is stable at 0.951. TrafficJam (0.766) is degraded by misclassification as Smooth (14\%), due to the ambiguous boundary between mild congestion and normal traffic flow. For VCR, LaneChange yields a notably low F1 of 0.217, with 78\% misclassified as Forward, as frame-level static processing is insufficient to capture gradual lateral movement. Turning also shows 18\% misclassification as Forward, reflecting the same temporal constraint. For DER, Anger (0.518) is the lowest-performing class due to limited samples (45) and suppressed facial expression during driving; Happiness (0.744) is degraded by misclassification as Peace (32\%). For DBR, Phone achieves the highest F1 (0.941) owing to visual distinctiveness, while Talking (0.594) and LookAround (0.654) suffer from subtle postural similarity to NormalDrive.

\begin{table}[h]
\centering
\caption{Per-class performance.}
\label{tab:per_class}
\small
\begin{tabular}{ll|ccc|r}
\toprule
Task & Class & Prec. & Rec. & F1 & N \\
\midrule
\multirow{3}{*}{TCR}
 & TrafficJam & 0.766 & 0.766 & 0.766 & 77 \\
 & Waiting & 0.900 & 0.947 & 0.923 & 133 \\
 & Smooth & 0.959 & 0.942 & 0.951 & 399 \\
\midrule
\multirow{5}{*}{VCR}
 & Parking & 0.899 & 0.922 & 0.910 & 154 \\
 & Turning & 0.661 & 0.732 & 0.695 & 56 \\
 & Backward & 0.769 & 0.769 & 0.769 & 39 \\
 & LaneChange & 0.357 & 0.156 & 0.217 & 32 \\
 & Forward & 0.860 & 0.881 & 0.870 & 328 \\
\midrule
\multirow{5}{*}{DER}
 & Anxiety & 0.610 & 0.726 & 0.663 & 84 \\
 & Peace & 0.842 & 0.848 & 0.845 & 363 \\
 & Weariness & 0.836 & 0.836 & 0.836 & 67 \\
 & Happiness & 0.889 & 0.640 & 0.744 & 50 \\
 & Anger & 0.550 & 0.489 & 0.518 & 45 \\
\midrule
\multirow{7}{*}{DBR}
 & Smoking & 0.792 & 0.679 & 0.731 & 28 \\
 & Phone & 0.956 & 0.926 & 0.941 & 94 \\
 & LookAround & 0.656 & 0.651 & 0.654 & 129 \\
 & DozingOff & 0.800 & 0.667 & 0.727 & 12 \\
 & NormalDrive & 0.792 & 0.831 & 0.811 & 248 \\
 & Talking & 0.559 & 0.633 & 0.594 & 30 \\
 & BodyMove & 0.726 & 0.662 & 0.692 & 68 \\
\bottomrule
\end{tabular}
\end{table}

\section{Detailed Analysis of the Psychological State Signal}
\label{sec:appendix_psi}

For TCR, the patterns of d4, d5, d6, d8, and d10 are reversed between TrafficJam and Waiting, consistent with cognitive science findings \citep{29} that driver internal states vary in response to qualitatively different stressors. For DER, Weariness shows the most extreme negative value on d16 among all classes, acquiring a uniquely distinct representation. Happiness shares a negative d7 with Anxiety but exhibits a distinct profile on d1, d2, and d15, suggesting that liveliness and anxiety share an arousal direction while being separated in valence. Anxiety and Anger both show positive values on d6, but Anger uniquely exhibits a high value on d4, consistent with the cognitive science insight that anxiety and anger differ in action tendency, avoidance for anxiety, approach for anger \citep{12}. For DBR, LookAround and Talking both exhibit strong negative values on d7 similarly to Anxiety, consistent with the shared characteristic of outward attention allocation. For VCR, Backward exhibits extreme negative values on d7 and d11, suggesting that the high attentional demands of reversing maneuvers are captured in $\boldsymbol{\psi}$.

Dimension d6 takes positive values for nearly all classes except Phone in DBR, suggesting it segregates the distinctive attentional state of smartphone use from all other states. Dimension d16 shows strong negative values for Weariness in DER and DozingOff in DBR, functioning complementarily to d7 in representing decreased arousal. These dimensional interpretations are post-hoc hypotheses and do not guarantee direct correspondence to specific psychological indicators; the meaning of each dimension is dynamically determined by the data and task structure. Nevertheless, the systematic task-label-dependent patterns, some consistent with cognitive science theory, provide evidence that CTPC acquires meaningful conditioning signals about driver internal states in a self-supervised manner.

%%%%%%%%%%%%%%%%%%%%%%%%%%%%%%%%%%%%%%%%%%%%%%%%%%%%%%%%%%%%

% \newpage
% \input{checklist.tex}

\end{document}